\documentclass{article}
\usepackage{spconf,amsmath,graphicx}
\usepackage{amsmath}
\usepackage{amssymb}
\usepackage{amsfonts}
\usepackage{bbm}
\usepackage{booktabs}
\usepackage[table]{xcolor}
\usepackage{tablefootnote}
\usepackage{url}
\usepackage{pifont}
\usepackage[ruled,vlined,linesnumbered]{algorithm2e}
\usepackage{multirow}

\newcommand{\ie}{{\it i.e.}}
\newcommand{\eg}{{\it e.g.}}

\newcommand{\argmin}{\mathop{\rm arg~min}\limits}

\usepackage[square,numbers]{natbib}
\newcommand{\Yang}[1]{\textcolor{black}{#1}} 

\newcommand{\YangProof}[1]{\textcolor{black}{#1}}

\title{Using panoramic videos for multi-person localization and tracking \\in a 3D panoramic coordinate}
\name{Fan Yang$^{\star \ddagger}$,
 Feiran Li$^{**}$,
 Yang Wu$^{\dagger}$,
 Sakriani Sakti$^{\star \ddagger}$,
and Satoshi Nakamura$^{\star \ddagger}$}
\address{
$^{\star}$Nara Institute of Science and Technology, Japan;
$^{**}$Osaka University, Japan;
$^{\dagger}$Kyoto University, Japan\\
$^{\ddagger}$RIKEN, Center for Advanced Intelligence Project, Japan}

\begin{document}
%
\maketitle
\begin{abstract}

3D panoramic multi-person localization and tracking are prominent in many applications, however, conventional methods using LiDAR equipment could be economically expensive and also computationally inefficient due to the processing of point cloud data. In this work, we propose an effective and efficient approach at a low cost. First, we utilize RGB panoramic videos instead of LiDAR data. Then, we transform human locations from a 2D panoramic image coordinate to a 3D panoramic camera coordinate using camera geometry and human bio-metric property (i.e., height). Finally, we generate 3D tracklets by associating human appearance and 3D trajectory. We verify the effectiveness of our method on three datasets including a new one built by us, in terms of 3D single-view multi-person localization, 3D single-view multi-person tracking, and 3D panoramic multi-person localization and tracking. Our code is available at \url{https://github.com/fandulu/MPLT}.


\end{abstract}
\begin{keywords}
3D localization, multi-target tracking, panoramic videos.
\end{keywords}

\vspace{-0.5em}
\section{Introduction}
\label{sec:intro}
\vspace{-0.5em}

In daily life, we understand surrounding visual scenes in \YangProof{3D}.
For instance, we usually decide how to interact with surrounding people by first localizing and tracking them in a 3D egocentric coordinate: when we are walking down the street, we plan our \YangProof{path} to avoid collisions by analyzing the trajectories of the surrounding people; when we see friends walking towards us, we might also walk to them and have a greeting. For applications (\eg, social robotics) that also require visual scene understanding, performing multi-person localization and tracking in a 3D coordinate is strongly desired.

\textbf{Typical single-view 3D coordinate \YangProof{localization}} methods fall into two categories: using depth sensors, or, using object size and camera geometry. Previous studies~\cite{osep2017combined,kollmitz2019deep} relied on depth sensors (\eg, LiDAR) and instance segmentation to obtain the target location in a 3D camera coordinate. In practice, however, the instance person segmentation algorithm is imperfect in crowd scenes, resulting in the assigning of incorrect locations to a target person. To some extent, these methods are more suitable for multiple vehicle tracking~\cite{cho2014multi}, since they are rigid objects with known shapes and the distance between them is generally larger. In contrast, other methods~\cite{scheidegger2018mono,sharma2018beyond} infer 3D camera-coordinate locations by object bounding box size and camera geometry. However, there is a scale variance between standing persons and sitting persons in terms of bounding box height. Moreover, when a person is near the camera, only the upper body can be observed. Consequently, simply taking the bounding box height as a reference is inappropriate. Recently, a study~\cite{bertoni2019monoloco} demonstrated that using the skeleton length can obtain more accurate locations than using bounding boxes. We embrace this \YangProof{idea into} our framework to obtain target locations in a single-view 3D camera coordinate. 

\textbf{Conventional multi-person tracking} takes two stages. The first detects each person by an object detector, and the second associates the cross-frame identities by considering their appearance similarity and trajectory trend~\cite{yu2016poi, wojke2017simple}. Most existing studies work on 2D/3D single-view~\cite{kollmitz2019deep, sharma2018beyond}, or, 2D panoramic multi-target tracking~\cite{ delforouzi2019deep,delforouzi2019polar}. However, previous works have limitations: the 2D tracking results could not be directly used in some applications (\ie, robotics) since the real-world coordinate is 3D; it is easy to lose the tracking target in a 3D narrow-angle-view coordinate since it only covers a part of the surrounding environment. Therefore, we propose 3D panoramic multi-target tracking to address the aforementioned issues.

\begin{figure*}[!h]
\centering
  \includegraphics[width=\linewidth]{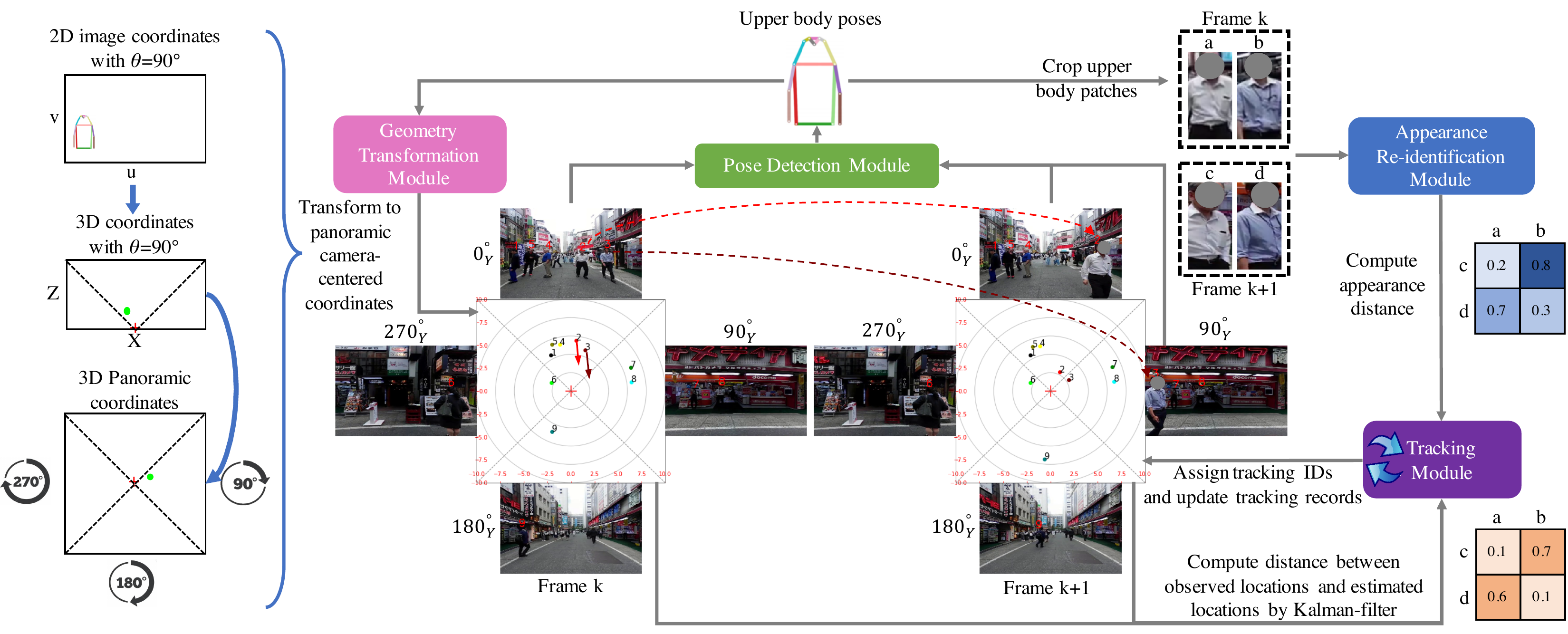}
  \vspace{-3mm}
  \caption{Our framework for 3D panoramic multi-person localization and tracking.}
  \label{fig:framework}
\end{figure*}

\textbf{We propose a novel framework for multi-person localization and tracking in a 3D panoramic coordinate with panoramic RGB videos}\footnote{Yang Wu is the corresponding author of this work.}. In our framework, 2D human poses are estimated for each person to obtain the 2D location and body height. Utilizing single-view intrinsic camera parameters, a person's 3D location can be approximated by assuming the body height is a constant. We further transform locations from a 3D single-view camera coordinate to a 3D panoramic coordinate using extrinsic camera parameters. Unlike in a 2D image coordinate, the real-scale location and motion are preserved in a 3D coordinate. As a benefit, it is easier to harness the power of Kalman filter to model human trajectories. To further address issues like occlusion and miss detection, we associate the appearance similarity and the trajectory trend together to approach multi-person tracking.

We annotated a Multi-person Panoramic Localization and Tracking (MPLT) dataset to evaluate our framework. We also compared our framework with others on the KITTI dataset~\cite{Geiger2012CVPR} and the 3D MOT dataset~\cite{andriluka2010monocular}, where only single-view 3D localization and tracking \YangProof{results} are provided.

\section{Methodology}
\label{sec:method}

Our framework includes four modules: Pose Detection Module, Geometry Transformation Module, Appearance Re-identification Module, and Tracking Module (see Fig.~\ref{fig:framework}). They work seamlessly together to achieve the target goal.

\subsection{Obtain 2D Person Poses}
Similar to previous work~\cite{bertoni2019monoloco}, we use off-the-shelf PifPaf~\cite{kreiss2019pifpaf} as our Pose Detection Module to estimate 2D human poses. Depends on the need, 2D person poses can be obtained either by a top-down approach or a bottom-up approach. In the former, an object detector (\eg, YOLO~\cite{redmon2016you}) is used to acquire the 2D bounding box for each person, and then PifPaf estimates 2D poses within each single bounding box. Alternatively, PifPaf can simultaneously estimate 2D poses for all persons and assign them to each person, which is a bottom-up approach. Compared with the top-down approach, the bottom-up approach is faster but less accurate.

\subsection{Coordinate Transformation}
We build a Geometry Transformation Module to map person locations from 2D image coordinates to a 3D panoramic coordinate. In our setting, four single-view cameras are used to capture panoramic videos. By removing the overlapping areas, we obtain four single-view images with a $90^{\circ}$ Horizontal Field of View at each frame. Following a clockwise path, we can assign each single-view image with view angle $\theta$, where $\theta \in \{0^{\circ}_{Y}, 90^{\circ}_{Y},180^{\circ}_{Y}, 270^{\circ}_{Y}\}$.

Let $[u_{\theta}, v_{\theta}]^{T}$ be a point in the 2D image coordinate and let $[X_{\theta}, Y_{\theta}, Z_{\theta}]^{T}$ be the corresponding point in the 3D camera coordinates of each single view. Then, we have 
\begin{equation}
\label{eq:intri}
\centering
\scalebox{0.8}{
\begin{math}
\begin{split}
\begin{bmatrix}
u_{\theta}\\ 
v_{\theta}\\ 
1\\
\end{bmatrix} = \mathbf{K}
\begin{bmatrix}
X_{\theta}\\ 
Y_{\theta}\\ 
Z_{\theta}\\ 
\end{bmatrix},\\
\end{split}
\end{math}}
\end{equation}
where $\mathbf{K}$ is the intrinsic matrix.

To transform locations from 3D camera coordinates to a 3D panoramic coordinate, we construct an extrinsic matrix:
\begin{equation}
\label{eq:extern}
\centering
\scalebox{0.8}{
\begin{math}
\begin{split}
\begin{bmatrix}
\mathbf{R}& \mathbf{t} \\ 
 \mathbf{0}^T& 1 
\end{bmatrix} 
\equiv [\mathbf{R}|\mathbf{t}]
\in \mathbb{R}^{4\times4} | \mathbf{R} \in SO(3), \mathbf{t} \in \mathbb{R}^{3},\\
\end{split}
\end{math}}
\end{equation}
where $SO(n)$ denotes a Special Orthogonal Group with dimension $n$; $\mathbf{0}$ indicates a zero vector; $\mathbf{R}$ and $\mathbf{t}$ are the 3D rotation matrix and the translation matrix. In our settings, all single-view coordinate centers are close to each other, so that $\mathbf{t}$ can be approximated by a zero vector and $\mathbf{R}$ only contain Y-axis rotation.  

Accordingly, for location $[X, Y, Z]^{T}$ in a 3D panoramic coordinate, the complete projection matrix can be defined:
\begin{equation}
\label{eq:tran}
\centering
\scalebox{0.8}{
\begin{math}
\begin{split}
\mathbf{P}_{\theta} = \mathbf{K}[\mathbf{R}(\theta)|\mathbf{t}],\\
\end{split}
\end{math}}
\end{equation}
and we have 
\begin{equation}
\label{eq:tran}
\centering
\scalebox{1}{
\begin{math}
\begin{split}
\begin{bmatrix} 
u_{\theta}\\ 
v_{\theta}\\ 
1
\end{bmatrix}
 = \mathbf{P}_{\theta}
\begin{bmatrix}
X\\ 
Y\\ 
Z\\ 
1
\end{bmatrix}.\\
\end{split}
\end{math}}
\end{equation}

At first glance, $[X, Y, Z]^{T}$ cannot be determined by $[u_{\theta},
v_{\theta}]^{T}$ in the above equation. However, we assume that real-world body height $H_{body}$ is a constant value. Since the corresponding body height in a 2D image coordinate (\ie, $h_{body}$) can be obtained by a pose estimator, for each person, the corresponding $X$ and $Z$ can be calculated by solving
\begin{equation}
\label{eq:loc}
\centering
\scalebox{0.8}{
\begin{math}
\begin{split}
\begin{bmatrix}
u_{\theta}\\ 
h_{body}\\ 
1
\end{bmatrix}= 
\mathbf{P}_{\theta}
\begin{bmatrix}
X\\ 
H_{body}\\ 
Z\\ 
1
\end{bmatrix}, \exists H_{body} \approx \text{constant}.\\
\end{split}
\end{math}}
\end{equation}

Hence, we can transform a target from a 2D single-view image coordinate to a 3D panoramic coordinate. Since \YangProof{most} real applications focus on the ground plane scenario, we treat $Y=0$ for all the persons in the 3D panoramic coordinate.

\subsection{Matching Cost}
The appearance of people can be utilized as an important tracking cue to alleviate the occlusion issue in tracking. Although existing works exploit the entire body appearance~\cite{yu2016poi, wojke2017simple}, we suppose that only using the upper body appearance can alleviate occlusion problems in crowd scenes. We further demonstrate this point in our experimental results of Table \ref{tab:MPLT}. Since 2D body poses are estimated in this work, the upper body image patches can be cropped accordingly.  We use an off-the-shelf model~\cite{Luo_2019_CVPR_Workshops} as our Appearance Re-identification Module. Given an upper-body image patch, it extracts the correspondent appearance embedding vector. 

In the tracking processes, appearance similarly is used to re-identify each person in the spatio-temporal domain. More specifically, the appearance cost between two consecutive frames is formulated as
\begin{equation}
\label{eq:app}
\scalebox{0.8}{
\begin{math}
\begin{split}
\mathbf{C}^{app}_{i,j} = 1-\frac{\mathbf{a}_{i}\mathbf{a}_{j}}{\left \| \mathbf{a}_{i} \right \| \left \| \mathbf{a}_{j} \right \|}, i\in \{1,\dots, N_i\}, j\in \{1,\dots, N_j\}
\end{split}
\end{math}}
\end{equation}
where $\mathbf{C}^{app}_{i,j}$ is the appearance cost of instance $i$ of the previous frame to instance $j$ of the current frame; $N_i$ and $N_j$ are the corresponding number of instances; $\mathbf{a}_{i}$ and $\mathbf{a}_{j}$ are the appearance embedding vectors with dimension $2048$.

Apart from the appearance cue, the trajectory trend is also a critical cue to track targets. With regard to previous works~\cite{yu2016poi, wojke2017simple}, Kalman filter~\cite{kalman1960new} is commonly used to model the trajectory trend. In contrast with modeling the trajectory trend in a 2D image coordinate, modeling it in a 3D coordinate can alleviate the position and motion distortions, which simplifies the procedure of applying Kalman filter to model trajectories. To be consistent with $\mathbf{C}^{app}$ at value range $0-1$, we apply an exponential kernel to calculate the distance between detected locations and Kalman filter estimated locations that are normalized by $H_{body}$. The trajectory cost between two consecutive frames is defined by
\begin{equation}
\label{eq:dis}
\scalebox{0.8}{
\begin{math}
\begin{split}
\mathbf{C}^{traj}_{i,j} = 1-\text{exp}\bigg( -\frac{(\hat{X}_{i}-X_{j})^{2}+(\hat{Z}_{i}-Z_{j})^{2}} {H_{body}^{2}} \bigg)
\end{split}
\end{math}}
\end{equation}
where $\mathbf{C}^{traj}_{i,j}$ is the trajectory cost of instance $i$ of the previous frame to instance $j$ of current frame. Additionally, $[\hat{X}_{i}, \hat{Z}_{i}]$ denotes the estimated location of instance $i$ at \textbf{current} frame by Kalman filter, while $\mathbf{L}_{j,:} = [X_{j}, Z_{j}]$ presents the detected location of instance $j$ at current frame, where $\mathbf{L}$ denotes the location values of all the detected instances.

We can simply associate $\mathbf{C}^{app}_{i,j}$ and $\mathbf{C}^{traj}_{i,j}$ by \YangProof{letting}

\begin{equation}
\label{eq:add}
\scalebox{0.8}{
\begin{math}
\begin{split}
\mathbf{C}_{i,j} = \mathbf{C}^{traj}_{i,j} + \mathbf{C}^{app}_{i,j},
\end{split}
\end{math}}
\end{equation}
where $\mathbf{C}_{i,j}$ is the associate cost of matching instance $i$ of the previous frame to instance $j$ of the current frame.
Then optimal assignment $\mathbf{M}^*$ is obtained by minimizing the total cost
\begin{equation}
\label{eq:hungarian}
\scalebox{0.8}{
\begin{math}
\begin{split}
\mathbf{M}^* = \argmin_{\mathbf{M}} \sum _{i}\sum _{j}\mathbf{C}_{i,j}\mathbf{M}_{i,j},
\end{split}
\end{math}}
\end{equation}
where $\mathbf{M}$ is a Boolean matrix. When row $i$ is assigned to column $j$, we have $\mathbf{M}_{i,j}=1$. Note that, each row can be assigned to at most one column and each column to at most one row. The optimization can be done by the Hungarian method.

\subsection{Multi-person Tracking}

\SetAlCapNameFnt{\footnotesize}
\SetAlCapFnt{\footnotesize}
\SetAlFnt{\footnotesize}
\begin{algorithm}[b!]
    \SetKwInOut{Input}{Input}
    \SetKwInOut{Output}{Output}
    \Input {$k$ (current tracked frame number), $\mathbf{C}$ (association cost matrix),  $\mathbf{L}$ (instance location matrix), $\mathbb{T}$ (active instance set), $\varepsilon$ (matching cost threshold)}
    \eIf{$k=1$}
    {
        Initialize active instance set $\mathbb{T} \leftarrow \varnothing $.\\
        \For{$j\gets1$ \KwTo $N_j$}{
        $\mathbb{T}_{j}[location] \leftarrow \mathbb{T}_{j}[location]\cup \{\mathbf{L}_{j,:}$\}.\\  $\mathbb{T}_{j}[lifespan] = 10$.}
        }
    {
        Obtain $\mathbf{M}^*$ by optimizing Eq.~\eqref{eq:hungarian} with the Hungarian method.\\
        Initialize the $N_j \times 1$ dimensional matching indicator vector $\mathbf{m} = \mathbf{0}$ for current frame $k$.\\
       
        \For{$i\gets1$ \KwTo $N_i$}{
            $\mathbb{T}_{i}[lifespan] = \mathbb{T}_{i}[lifespan] - 1$. \\
            \For{$j\gets1$ \KwTo $N_j$}{
        \If{$\mathbf{M}^*_{i,j}=1$ and $\mathbf{C}_{i,j}< \varepsilon$}
        {
        $\mathbb{T}_{i}[location] \leftarrow \mathbb{T}_{i}[location]\cup \{\mathbf{L}_{j,:}$\}.\\  $\mathbb{T}_{i}[lifespan] = 10$.\\
        $\mathbf{m}_j = 1$.        
        }
        }
        }
        \For{$j\gets1$ \KwTo $N_j$}{
            \If{$\mathbf{m}_j = 0$ (\ie, instance $j$ is unmatched)}
            {
            Add one more active instance to $\mathbb{T}$: \\
            $\mathbb{T}_{|\mathbb{T}|+1}[location] \leftarrow \mathbb{T}_{|\mathbb{T}|+1}[location]\cup \{\mathbf{L}_{j,:}\}$.\\
            $\mathbb{T}_{|\mathbb{T}|+1}[lifespan] = 10$.
            }
        }
    }
    \If {$\mathbb{T}_{i}[lifespan]=0$}
     {Remove $\mathbb{T}_{i}$ from $\mathbb{T}$.}
     \For{$l\gets1$ \KwTo $|\mathbb{T}|$}{
     Update Kalman filter with $\mathbb{T}_{l}[location]$.\\
      $\mathbb{T}_{l}[location\_estimated] = [\hat{X}_l, \hat{Z}_l]$, estimated using the updated Kalman filter.}
    \Output{$\mathbb{T}$}
    \caption{\textbf{Tracking algorithm}}
    \label{algorithm1}
\end{algorithm}

In the Tracking Module, we create a tracking set $\mathbb{T}$ to store and update tracked instances. At \Yang{the $k$-th tracked} frame, we obtain a set of 3D panoramic locations $\mathbf{L}$ by Eq.~\eqref{eq:loc} and the cost matrix \Yang{$\mathbf{C}$} by Eq.~\eqref{eq:add}. In the first frame, all the observed locations are assigned to a tracking set. After that, the across-frame connections are determined by $\mathbf{M}_{i,j}$ and $\Yang{\mathbf{C}}_{i,j}$. When $\mathbf{M}_{i,j}=1$ and $\Yang{\mathbf{C}}_{i,j}$ is smaller than a threshold $\varepsilon$, the instance $i$ of frame $k-1$ is likely to be the instance $j$ of frame $k$. However, across-frame instances may not always be perfectly matched. For unmatched instance $j$, we assign it to \Yang{$\mathbb{T}$} as a new instance. For unmatched instance $i$, which is already recorded in \Yang{$\mathbb{T}$}, we reduce its lifespan by $1$. While new instances come into the tracking area, old instances may also leave. Therefore, we delete unseen instances in the tracking set after $10$ frames. We summarize this process in Algorithm~\ref{algorithm1}.

\vspace{-7px}
\section{Experiments}
\label{sec:exp}

\noindent \textbf{Experimental Datasets.}
We annotate a Multi-person Panoramic Localization and Tracking (MPLT) Dataset to enable model evaluation on 3D panoramic multi-person localization and tracking. It represents real-world scenarios and contains a crowd of people in each frame. And, over 1.8K frames and densely annotated 3D trajectories are included. For comparison with related works, we also evaluate our framework on the KITTI~\cite{Geiger2012CVPR} and 3D MOT~\cite{andriluka2010monocular} datasets. The properties of three experimental datasets are listed as follows:
\vspace{-17px}
\begin{table}[!h]
\centering
\caption{Properties of experimental datasets.}
\scalebox{0.76}{
\begin{tabular}{lccc}
\toprule
\textbf{Dataset} &\textbf{3D single-view} &\textbf{3D single-view} &\textbf{3D panoramic}\\ 
&\textbf{localization} &\textbf{localization\&} &\textbf{localization\&}\\ 
& & \textbf{tracking}& \textbf{tracking}\\
\midrule
KITTI~\cite{Geiger2012CVPR} & \ding{52} & & \\
3D MOT~\cite{andriluka2010monocular}&  &\ding{52} & \\
MPLT &  & & \ding{52}\\
\bottomrule
\end{tabular}}
\end{table}

\noindent \textbf{Experimental Setup.}
Since off-the-shelf pose detector and appearance extractor are applied, we do not train any \YangProof{models in this work}. For the KITTI and MPLT datasets, we apply the bottom-up pose estimation approach. For 3D MOT dataset, we apply the top-down pose estimation with the given public bounding boxes. Based on the properties of each dataset, we evaluate the model performance from different perspectives. 

\noindent \textbf{Experimental Results.}
In Table ~\ref{tab:KITTI}, we report the localization precision under three thresholds for the KITTI Dataset. It shows \YangProof{good} generalization property of our method. Although without any training on KITTI Dataset, \YangProof{its performance} can reach the second place in terms of 3D single-view localization without any extra training. In Table ~\ref{tab:MOT_3D}, we compare our framework with others on the 3D MOT Benchmark\footnote{\url{https://motchallenge.net/results/3D_MOT_2015/}}, which targets at 3D single-view localization and tracking. We achieve the state-of-the-art performance (\ie, $1^{st}$ place of the public leaderboard) on the dominant criterion (\ie, MOTA \cite{ristani2016performance}), which outperforms the second place method by $1.5$. For our proposal dataset MPLT, we list the performance of our framework and make it as a baseline (see Table ~\ref{tab:MPLT}). Furthermore, we also show, due to the occlusion, selecting the whole body appearance may impair the model performance. The qualitative evaluation results are available on our project page \footnote{\url{https://github.com/fandulu/MPLT}}.

\vspace{-17px}
\begin{table}[!h]
\centering
\caption{Monocular-camera-based localization precision on KITTI Dataset. If distance from predicted locations to ground-truth location is within a threshold, it is correctly predicted.}
\scalebox{0.76}{
\begin{tabular}{lccc}
\toprule
 \multirow{2}{*}{\textbf{Methods}} & \multicolumn{3}{c}{\textbf{Localization precision by threshold}}\\
 \cline{2-4}
&$<0.5 m$ &$<1.0 m$ &$<2.0 m$\\
\cline{1-4}
Mono3D~\cite{chen2016monocular}(training on KITTI) &13.2\% &23.3\% &39.0\%\\
SAMono\cite{xu2018structured}(training on KITTI)&19.8\% &33.9\% &48.5\%\\
MonoDepth~\cite{godard2017unsupervised}(training on KITTI)&20.6\% &35.4\% &50.7\%\\ 
MonoLoco~\cite{bertoni2019monoloco} (training on KITTI)&\textbf{29.0}\% &\textbf{49.6}\% &\textbf{71.2}\%\\\hline
Ours (w/o KITTI) &\underline{22.0\%} &\underline{39.4\%} &\underline{63.5\%}\\
\bottomrule
\end{tabular}}
\label{tab:KITTI}
\end{table}

\vspace{-23px}
\begin{table}[!h]
\caption{3D MOT Benchmark. $\uparrow$($\downarrow$) indicates that the larger(smaller) the value is, the better the performance. Multiple Object Tracking Accuracy (MOTA) is the dominant criterion. The details of the evaluation metrics were previously explained in \cite{ristani2016performance}.}
\label{tab:MOT_3D}
\centering
\scalebox{0.73}{
\begin{tabular}{lcccccc}
\toprule
\textbf{Methods}  &\cellcolor{gray!20}\textbf{MOTA}$\uparrow$	 &\textbf{MT}$\uparrow$ &\textbf{ML}$\downarrow$ &\textbf{FP}$\downarrow$&\textbf{FN}$\downarrow$\\ 
\midrule
AMIR3D~\cite{sadeghian2017tracking} &\cellcolor{gray!20}25.0	&3.0\%	&27.6\%	&2,038	&9,084	\\ 
MCFPHD~\cite{wojke2016global} &\cellcolor{gray!20}39.9 &25.7\%	&\textbf{16.8\%}	&3,029	&6,700 \\ 
GPDBN~\cite{klinger2017probabilistic} &\cellcolor{gray!20}49.8	&25.7\%	&17.2\%	&\textbf{1,813}	&6,300\\ 
MOANA~\cite{tang2019moana} &\cellcolor{gray!20}52.7	&28.4\%	 &22.0\%	&2,226	&5,551 \\ \hline
Ours & \cellcolor{gray!20}\textbf{54.2}	&\textbf{30.6\%}	&20.9\%	&2,385	&\textbf{4,930} \\ 
\bottomrule
\end{tabular}}
\end{table}

\vspace{-23px}
\begin{table}[!h]
\centering
\caption{Performance of our framework on MPLT dataset. We evaluate localization and tracking performance within $10$ m of the coordinate center. }
\scalebox{0.76}{
\begin{tabular}{llc}
\toprule
\textbf{Appearance Selection}
&\textbf{Threshold}  &\cellcolor{gray!20}\textbf{MOTA}$\uparrow$	\\ 
\midrule
Whole body&$<0.5 m$ & \cellcolor{gray!20}62.4 \\
Whole body&$<1.0 m$ & \cellcolor{gray!20}70.2 \\ \hline
Upper body&$<0.5 m$ & \cellcolor{gray!20}\textbf{65.2}  \\
Upper body&$<1.0 m$ & \cellcolor{gray!20}\textbf{74.9} \\ 
\bottomrule
\end{tabular}}
\label{tab:MPLT}
\end{table}

\section{conclusion}
We proposed a simple yet effective solution for 3D panoramic multi-person localization and tracking with panoramic videos. On two existing datasets, the effectiveness of our method is demonstrated by the promising performance. Meanwhile, a strong baseline is offered for our new benchmark dataset. Since our method can faithfully keep the realistic locations and motions for tracking targets in a 3D panoramic coordinate, it can help human-related video understanding applications. \YangProof{As} future work, we \YangProof{plan to} integrate our framework with a previous work~\cite{yang2019framework} for automatically detecting human activities in a 3D panoramic coordinate.

\noindent \textbf{ACKNOWLEDGEMENTS.}
Part of this work was supported by JSPS KAKENHI Grant Numbers JP17H06101 and JP17K00237\YangProof{, and a MSRA Collaborative Research 2019 Grant by Microsoft Research Asia.}


\bibliographystyle{IEEEbib}
{\small\bibliography{refs}}

\end{document}